\documentclass{article}

\usepackage[preprint]{neurips_2026}

\usepackage{graphicx}
\graphicspath{{figures/}}
\usepackage[utf8]{inputenc} % allow utf-8 input
\usepackage[T1]{fontenc}    % use 8-bit T1 fonts
\usepackage{hyperref}       % hyperlinks
\usepackage{url}            % simple URL typesetting
\usepackage{booktabs}       % professional-quality tables
\usepackage{amsfonts}       % blackboard math symbols
\usepackage{nicefrac}       % compact symbols for 1/2, etc.
\usepackage{microtype}      % microtypography
\usepackage{xcolor}         % colors
\usepackage{enumitem}       % customized list spacing
\usepackage[table]{xcolor}
\definecolor{lightblue}{RGB}{225,238,250}
\definecolor{improveblue}{RGB}{30,90,200}
\usepackage{multirow}
\usepackage{caption}
\usepackage{wrapfig}
\usepackage{placeins}
\usepackage{fancyvrb}
\usepackage{array}
\newcolumntype{L}[1]{>{\raggedright\arraybackslash}p{#1}}
\newcolumntype{C}[1]{>{\centering\arraybackslash}p{#1}}
\newcommand{\gain}[1]{\textcolor{improveblue}{\footnotesize($\uparrow$#1)}}
\usepackage[most]{tcolorbox}
\usepackage{xspace}
\newcommand{\method}{DataEvolver\xspace}

\usepackage{titlesec}

\titleformat*{\section}{\large\bfseries\color{seedblue}}
\titleformat*{\subsection}{\normalsize\bfseries\color{seedblue}}
\titleformat*{\subsubsection}{\normalsize\bfseries\color{seedblue}}

\definecolor{seedblue}{HTML}{2E5AA8}

\definecolor{deblueA}{HTML}{0A3A7A}
\definecolor{deblueB}{HTML}{1359B8}
\definecolor{decyanA}{HTML}{248CC8}
\definecolor{detealA}{HTML}{2AA6A6}
\definecolor{projectpink}{HTML}{E91E63}

\makeatletter
\renewcommand{\@notice}{}

\renewcommand{\@toptitlebar}{%
  {\color{seedblue}\hrule height 2.5pt}%
  \vskip 2mm
  \vskip -\parskip
}

\renewcommand{\@bottomtitlebar}{%
  \vskip 5mm
  \vskip -\parskip
  {\color{seedblue}\hrule height 1pt}%
  \vskip 2mm
}
\makeatother

\renewenvironment{abstract}{%
  \vskip 0.075in
  \centerline{\large\bfseries\textcolor{seedblue}{Abstract}}
  \vspace{0.5ex}
  \begin{quote}
}{%
  \par\end{quote}\vskip 1ex
}

\DeclareRobustCommand{\DataEvolverGradient}{%
\textcolor{deblueA}{D}%
\textcolor{deblueA}{a}%
\textcolor{deblueB}{t}%
\textcolor{deblueB}{a}%
\textcolor{decyanA}{E}%
\textcolor{decyanA}{v}%
\textcolor{detealA}{o}%
\textcolor{detealA}{l}%
\textcolor{detealA}{v}%
\textcolor{detealA}{e}%
\textcolor{detealA}{r}%
}

\DeclareRobustCommand{\DataEvolverLogo}{%
  \raisebox{-0.25em}{%
    \includegraphics[
      height=1.95em
    ]{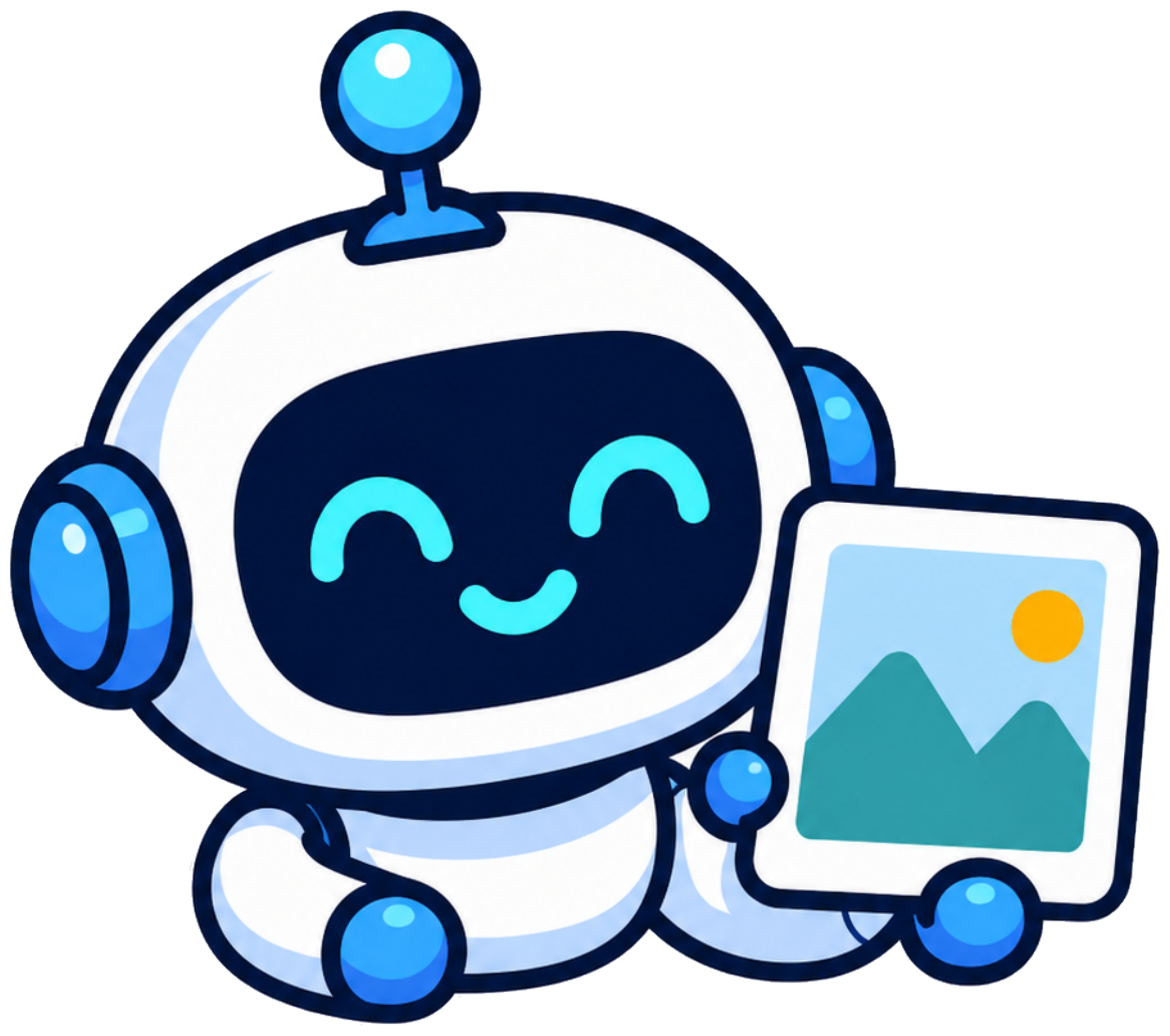}%
  }%
}

\setlist[itemize]{leftmargin=1.2em,itemsep=0.25em,topsep=0.25em,parsep=0pt,partopsep=0pt}

\title{%
\texorpdfstring{\DataEvolverLogo\hspace{0.35em}\DataEvolverGradient}{\method}: Self-Evolving Multi-Agent Data Construction for Text-Rich Image Generation
}

\author{%
\normalfont
{\bfseries
\textsuperscript{1,2}Siyu Yan\thanks{Equal contribution.}
\quad
\textsuperscript{1}Yizhen Gao\footnotemark[1]
\quad
\textsuperscript{1}Yilin Wang
\quad
\textsuperscript{1}Dongxing Mao
\quad
\textsuperscript{1}Alex Jinpeng Wang\thanks{Corresponding author.}
}\\[0.35em]
\textsuperscript{1}Central South University
\quad
\textsuperscript{2}The Hong Kong University of Science and Technology\\[0.45em]
Project Page: \href{https://github.com/CSU-JPG/DataEvolver}{\textcolor{projectpink}{\ttfamily https://github.com/CSU-JPG/DataEvolver}}
}

\begin{document}

\maketitle

\begin{abstract}
Text-rich image generation is one of the most challenging settings in image generation, since models must simultaneously produce visually realistic images and render legible, semantically aligned, and layout-consistent text. Existing data pipelines usually follow a static crawl--filter--freeze paradigm. They collect candidate samples, filter them once, and freeze the accepted data for training. However, rejected samples are usually discarded, although they often contain useful failure signals such as OCR errors and semantic mismatches. As a result, later construction rounds may repeat the same failure modes. To address these limitations, we propose \textbf{\method}, a self-evolving multi-agent framework for text-rich image data construction. \method treats data construction as feedback-driven construction policy evolution. A Retriever collects candidate samples, a Verifier assigns quality scores and rejection causes, a Critic summarizes round-level feedback into semantic feedback, and a Generator completes under-covered regions through targeted synthesis. The updated feedback memory then guides the next construction round.
Experiments on text-rich image generation benchmarks show that \method produces more useful training data than fixed-dataset baselines under matched data budgets.
At the 0.75M scale on PixArt-$\alpha$, \method improves OCR-F1 over the strongest baseline by \textbf{85.3\%} on TextScenesHQ and \textbf{35.3\%} on LongTextBench.
The improvements are consistent across both evaluated benchmarks and also transfer to Show-o2, indicating that the benefit of \method is not tied to a single downstream generator.
These results suggest that rejected samples can provide actionable feedback for improving text-rich image data construction.

\end{abstract}

\section{Introduction}
Modern text-to-image generation has advanced rapidly with diffusion and transformer-based architectures~\citep{glide,dalle2,imagen,ldm,pixartalpha,showo2}. However, model progress is increasingly constrained by the data construction pipelines that supply training supervision. This limitation is especially clear in text-rich image generation. Training samples must satisfy visual fidelity, text legibility, semantic alignment, and layout coherence at the same time. Existing large-scale multimodal datasets have enabled substantial progress~\citep{cc12m,laion5b,coyo700m,datacomp,mmc4,obelics,textdiffuser,anytext,textdiffuser2,textatlas5m}, but constructing high-quality text-rich image data at scale remains difficult. Failure cases are diverse and recurring, and they often concentrate in specific topics, layouts, fonts, or text patterns~\citep{glyphdraw,glyphcontrol,artist}.

Most existing data construction pipelines follow a static crawl--filter--freeze paradigm~\citep{laion5b, datacomp, textdiffuser, anytext, textatlas5m}.
They collect candidate samples, apply fixed filtering rules, and then freeze the accepted data into a training set. While effective for large-scale collection, this paradigm discards useful construction-time feedback. Rejected samples can reveal unreliable retrieval queries, under-covered semantic regions, low-quality retrieval patterns, and recurring generation or recognition failures. Yet such signals are usually treated only as filtering outcomes, rather than as guidance for improving later construction rounds.

\begin{figure}[t]
    \centering
    \includegraphics[width=1\linewidth]{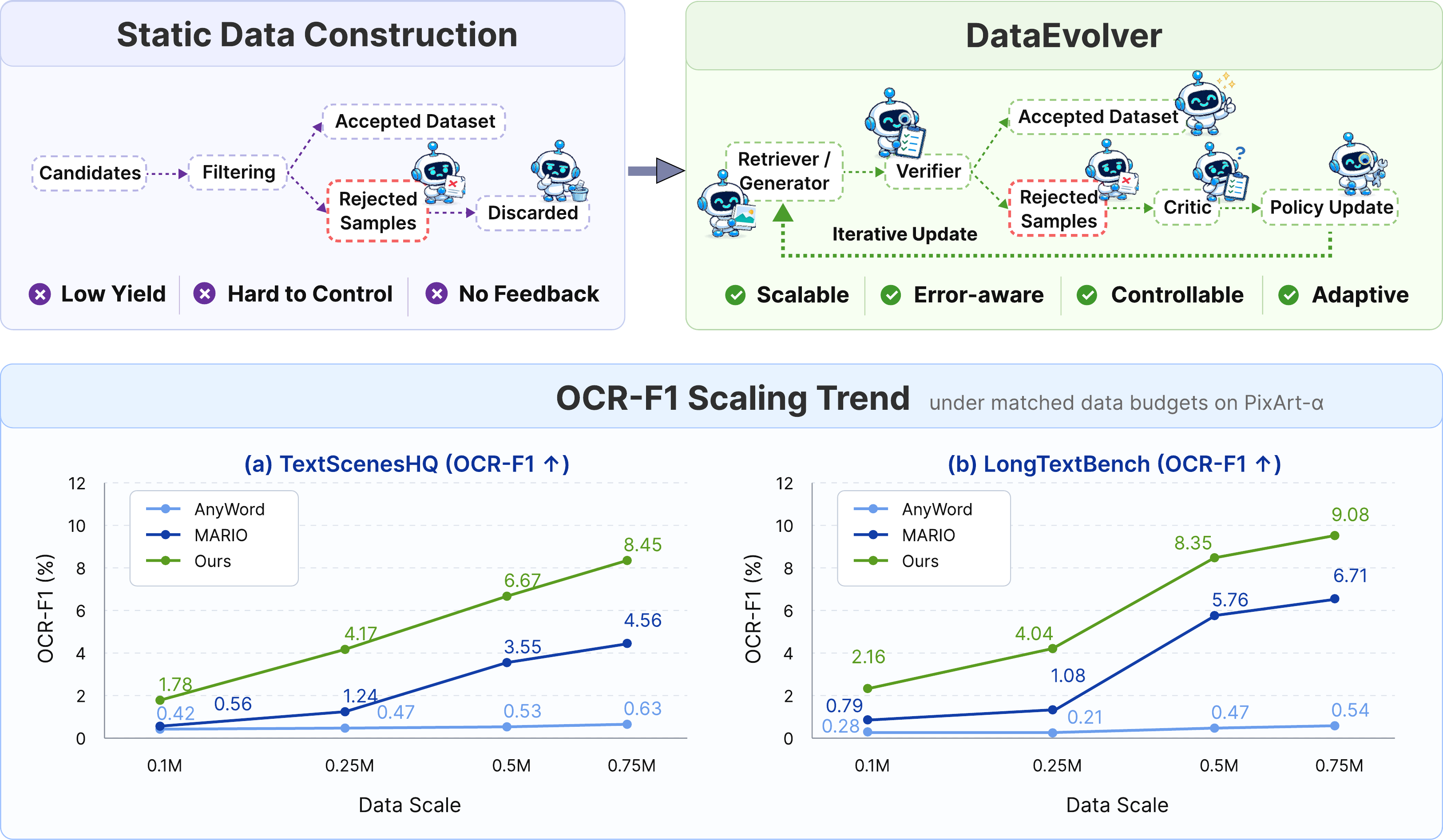}
    \caption{
    \textbf{Turning Rejected Samples into Feedback.}
    The upper panel contrasts static data construction, which discards rejected samples, with \method, which converts rejection patterns into feedback for iterative policy updates.
    The bottom panel shows OCR-F1 scaling trends on PixArt-$\alpha$~\citep{pixartalpha} under matched data budgets, comparing \method with AnyWord~\citep{anytext} and MARIO~\citep{textdiffuser}.
    }
    \label{fig:overview}
\end{figure}

This observation motivates a different view of dataset construction. Rather than treating failure cases as unusable byproducts of filtering, we treat them as practical diagnostic evidence for improving the construction process.
Rejected samples can indicate which retrieval queries repeatedly produce duplicates and which topics remain under-covered. The key challenge is therefore not only to collect more samples, but to convert these construction-time failures into explicit feedback that guides retrieval planning and targeted completion in later rounds.

In this work, we propose \textbf{\method}, a self-evolving multi-agent framework for autonomous text-rich image data construction, as illustrated in Figure~\ref{fig:overview}. \method consists of four cooperative agents. The \textbf{Retriever} discovers candidate samples. The \textbf{Verifier} converts candidates into pass sets and rejection statistics. The \textbf{Critic} distills verification feedback and rejection patterns into natural-language \emph{semantic feedback}. The \textbf{Generator} performs targeted completion for under-covered regions. Together, these agents form a closed loop that turns construction feedback into reusable guidance for subsequent rounds.

We instantiate \method for text-rich image generation and evaluate the constructed data under matched data budgets. Using the same downstream generator for fine-tuning, we compare \method with strong public data sources and ablated variants on TextScenesHQ and LongTextBench. Experimental results show that \method consistently improves OCR-oriented metrics that directly reflect text rendering quality, while maintaining competitive semantic alignment and visual quality. Process-level analyses further show that the evolving construction policy tends to produce samples with higher OCR confidence, suggesting that part of the improvement emerges during data construction itself rather than only after downstream model training.

Our contributions are summarized as follows:
\begin{itemize}[leftmargin=1.2em,labelsep=0.5em,itemsep=0.3em,topsep=0.2em,parsep=0pt,partopsep=0pt]
    \item \textbf{A failure-aware view of text-rich image data construction.}
We reformulate text-rich image data construction as construction policy evolution, where rejected samples and failure cases are treated as reusable feedback for improving the construction process rather than as discarded filtering byproducts.

    \item \textbf{A self-evolving multi-agent construction framework.}
    We introduce \textbf{\method}, a closed-loop framework that transforms round-level verification outcomes into reusable semantic feedback for subsequent construction rounds.

    \item \textbf{Multi-scale evaluation for text-rich image generation.}
    We evaluate \method under matched data budgets on TextScenesHQ and LongTextBench. At the 0.75M scale on PixArt-$\alpha$, \method improves OCR-F1 over the strongest baseline by \textbf{85.3\%} on TextScenesHQ and \textbf{35.3\%} on LongTextBench. Ablation results further show that removing the Critic or Generator consistently degrades OCR-F1, which validates both feedback-based policy revision and targeted completion.
\end{itemize}
\section{Related Work}

\paragraph{Multimodal Data Construction.}
Large-scale vision-language datasets are commonly built through web crawling, offline filtering, and deduplication. Representative datasets such as CC12M~\citep{cc12m}, LAION-5B~\citep{laion5b}, COYO-700M~\citep{coyo700m}, and DataComp~\citep{datacomp} have shown that data sources, filtering criteria, and dataset composition strongly affect downstream model performance. Beyond paired image-text data, MMC4~\citep{mmc4} and OBELICS~\citep{obelics} construct interleaved image-text corpora from web documents, while JourneyDB~\citep{journeydb} collects generated image-prompt pairs. For text-rich image generation, recent datasets and methods further emphasize text legibility, layout structure, and dense or long textual content. TextDiffuser introduces MARIO-10M~\citep{textdiffuser}, AnyText builds AnyWord-3M~\citep{anytext}, and TextDiffuser-2~\citep{textdiffuser2} and TextAtlas5M~\citep{textatlas5m} extend data construction toward long-text, dense-text, and structured-text scenarios. Related methods such as GlyphDraw~\citep{glyphdraw}, GlyphControl~\citep{glyphcontrol}, and ARTIST~\citep{artist} also highlight the importance of explicit modeling for character legibility, text placement, and layout consistency. Recent LLM-based data construction methods, such as AgentInstruct~\citep{agentinstruct}, MMInstruct~\citep{mminstruct}, and Oasis~\citep{oasis}, explore agentic or multi-model data synthesis. However, most existing pipelines still use feedback mainly for filtering or quality control. In contrast, \method uses construction-time failures to revise the construction policy itself.

\paragraph{Feedback-Driven Self-Improvement.}
Feedback-driven learning and self-improvement have been widely studied for language models and agents. RLHF~\citep{rlhf}, PPO~\citep{ppo}, and DPO~\citep{dpo} optimize model behavior with human feedback or preference data, while Constitutional AI~\citep{constitutionalai} and RLAIF~\citep{rlaif} reduce the reliance on direct human annotation by using AI-generated feedback or rule-based principles. Beyond preference optimization, iterative refinement methods use model-generated rationales, external critiques, execution feedback, or memory to improve later outputs. Representative examples include STaR~\citep{star}, ReST~\citep{rest}, Self-Refine~\citep{selfrefine}, Reflexion~\citep{reflexion}, and CRITIC~\citep{critic}. Recent agent systems further show that reflection, tool use, and memory can improve long-horizon behavior~\citep{react,toolformer,voyager}. These works demonstrate the value of feedback loops, but they mainly improve model outputs, reasoning traces, or task policies. \method instead applies feedback-driven improvement to data construction, converting rejected samples into semantic policy feedback for later construction rounds.
\section{Preliminaries and Problem Setup}
\label{sec:preliminaries}

\subsection{Autonomous Data Construction as Iterative Policy Refinement}
\label{sec:policy_refinement}

We study autonomous multimodal data construction for text-rich image generation.
Given a target domain consisting of topics and subtopics, our goal is to build a high-quality dataset
that can effectively support downstream generation models. Unlike conventional pipelines that treat
dataset construction as a one-shot process of crawling, filtering, and storage, we formulate it as an
iterative policy refinement process in which the data construction strategy is updated according to
observed construction-time feedback.

At round $t$, the system is governed by a data construction policy
\[
\pi_t = (\mathcal{Q}_t, \mathcal{P}_t, \mathcal{E}_t),
\]
where $\mathcal{Q}_t$ denotes the retrieval policy, $\mathcal{P}_t$ denotes the generation prompt policy, and $\mathcal{E}_t$ denotes an experience library that stores high-value feedback and policy traces from previous rounds. The Verifier configuration and verification criteria are kept fixed in our reported experiments, and are not treated as optimized policy variables. Under policy $\pi_t$, the system first produces a raw candidate set
\[
\mathcal{I}_{\text{raw}}^t = \{i_1, i_2, \dots, i_n\},
\]
which is then evaluated by the fixed verification procedure to obtain a filtered pass set
\[
\mathcal{I}_{\text{pass}}^t \subseteq \mathcal{I}_{\text{raw}}^t.
\]

From this perspective, data construction is no longer a static preprocessing pipeline, but a closed-loop
process of \emph{policy execution, feedback collection, and policy revision}. In each round, the current
policy determines how candidate samples are retrieved or generated, while the resulting feedback is used
to revise the next-round policy. Therefore, the target of improvement is not merely a fixed batch of
collected data, but the construction process that governs how the dataset is produced.

\subsection{Round-Level Feedback Signals}
\label{sec:feedback_signals}

To support iterative policy refinement, we summarize the outcome of each construction round with a
multi-dimensional feedback representation. Rather than relying on a single binary pass/fail signal, we
capture both the overall quality of accepted samples and the dominant failure modes of rejected ones.
Specifically, the feedback signal of round $t$ is represented as
\[
s_t = (\rho_t, \bar{O}_t, \bar{C}_t, \bar{I}_t, \mathbf{R}_t),
\]
where $\rho_t$ is the pass rate, $\bar{O}_t$ is the mean OCR quality, $\bar{C}_t$ is the mean semantic
consistency score, $\bar{I}_t$ is the mean image quality score, and $\mathbf{R}_t$ is the rejection-cause
vector that records the distribution of major failure types such as blur, OCR failure, semantic mismatch,
or layout corruption.

These feedback signals serve two purposes. First, they measure the effectiveness of the current policy in
producing useful data. Second, they provide fine-grained information about \emph{why} a round succeeds
or fails, which is critical for updating subsequent retrieval queries, generation prompts, and experience memory.

Our objective is to progressively improve the quality and utility of the constructed dataset through
construction-time feedback. In practice, we update the controllable variables of the data pipeline,
including retrieval queries, generation prompts, and experience memory. These
updates are performed through iterative policy refinement rather than gradient-based optimization or
model-parameter training.

At round $t$, the Critic summarizes the current feedback signal into semantic feedback
\[
\mathcal{F}_t
=
\mathrm{Critic}(s_t, s_{t-1}, \mathcal{E}_{t-1}),
\]
which is then used to revise the next-round construction policy
\[
\pi_{t+1}
=
\mathrm{Compose}(\pi_t, \mathcal{F}_t, \mathcal{E}_t).
\]

Unlike gradient-based optimization, this refinement operates entirely in natural-language policy space.
The revised policy modifies retrieval queries, generation prompts, and experience memory without updating model parameters.

\begin{figure}[t]
    \centering
    \includegraphics[width=1\linewidth]{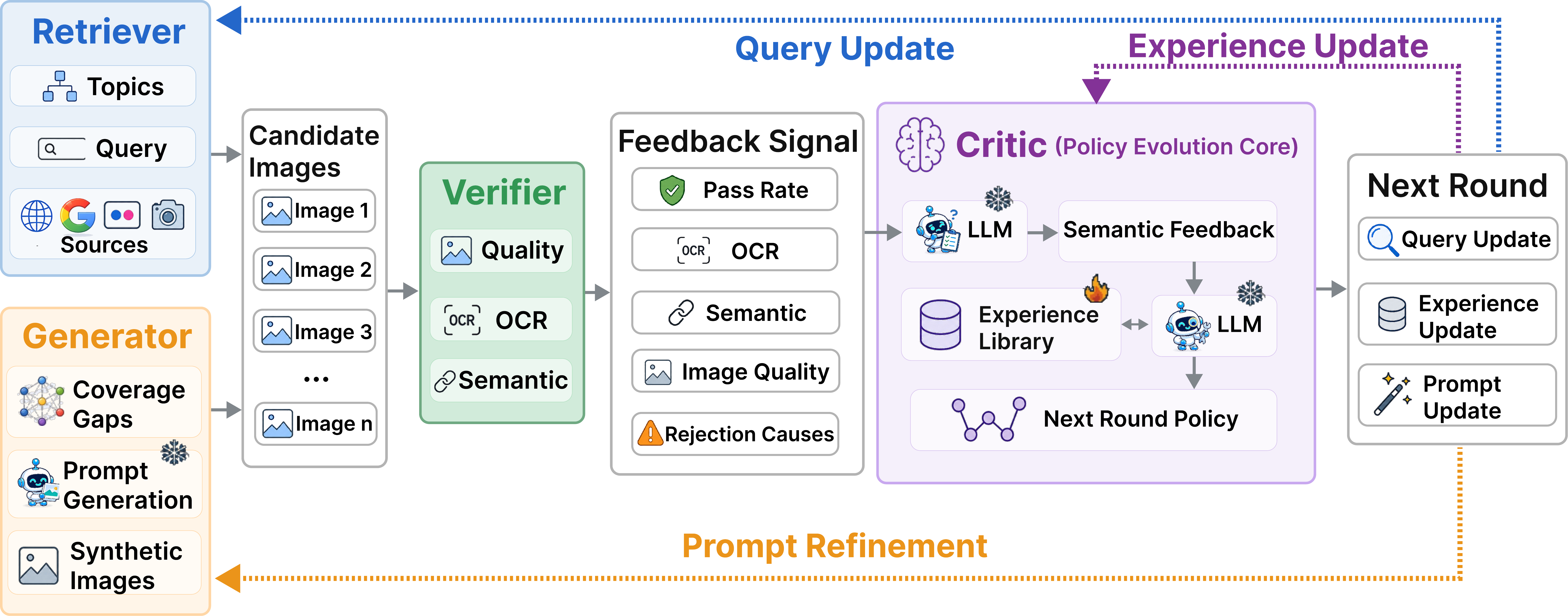}
    \caption{
    \textbf{Illustration of the DataEvolver framework.}
    DataEvolver constructs text-rich image data through a closed loop of retrieval, fixed verification, critic-guided policy update, and targeted synthesis.
    Rejected samples are converted into semantic feedback for query diversification, prompt refinement, and experience-memory update.
    }
    \label{fig:pipeline}
\end{figure}
\section{\method}
\label{sec:dataevolver}

As illustrated in Figure~\ref{fig:pipeline}, \method constructs text-rich image data through a closed loop with four agents.
The \textbf{Retriever} builds a candidate image pool from topic requirements.
The \textbf{Verifier} evaluates candidates and produces both pass sets and rejection causes.
The \textbf{Critic} summarizes round-level feedback for policy revision.
The \textbf{Generator} fills under-covered regions through targeted synthesis.

\subsection{Retriever: Queries to Candidate Pool}
\label{sec:retriever_candidates}

The first stage of \method transforms topic requirements into a candidate image pool.
Rather than relying on fixed keywords, the Retriever operates under a policy-conditioned query space that adapts over time according to previous feedback.
Given the current retrieval policy, it generates structured query templates for different topics and subtopics.
These templates may incorporate topic keywords, subtopic descriptors, semantic expansions, and source-level constraints, so that retrieval is guided not only by relevance, but also by coverage and diversity.

After query planning, the Retriever gathers candidate images from external sources and associates each sample with metadata such as topic, subtopic, query template, and source information.
The final output of this stage is a topic-aware candidate pool that serves as the input to the Verifier.

\subsection{Verifier: Candidates to Pass Set and Feedback}
\label{sec:verifier_passset}

The candidate pool cannot be directly admitted to the final dataset.
The role of the Verifier is to transform raw candidates into a pass set through multi-dimensional quality assessment and structured failure attribution.

For each candidate sample, OCR is first applied to extract text signals.
The Verifier then removes near-duplicate samples using perceptual hashing and evaluates three primary aspects: perceptual image quality, text recognition quality, and semantic consistency with the intended topic or subtopic.
Only samples that satisfy the predefined verification criteria are admitted to the pass set.
Beyond binary acceptance, the Verifier also summarizes the current round into the feedback signal defined in Section~\ref{sec:feedback_signals}.
In addition to aggregate quality statistics, it records the dominant failure modes among rejected samples, such as duplicate samples, blur, OCR failure, semantic mismatch, or layout corruption.

\subsection{Critic: Feedback to Policy Updates}
\label{sec:critic_policy_update}

The feedback signal captures the quality of a construction round, but numerical statistics alone are insufficient to directly revise the next construction policy.
The Critic addresses this problem by converting round-level feedback into natural-language update signals for query diversification, prompt refinement, and experience-memory updates.
The resulting semantic feedback summarizes dominant failure patterns, under-covered topics, duplicate-heavy queries, ineffective query templates, and directions for query or prompt refinement.

\begin{figure}[t]
    \centering
    \includegraphics[width=1\linewidth]{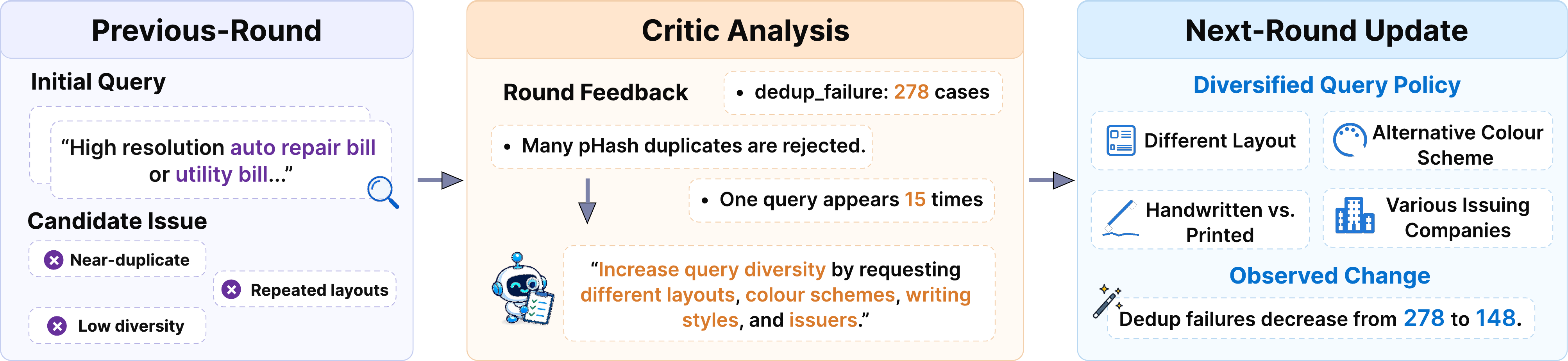}
    \caption{
    \textbf{Example of Critic-guided policy update.}
    Given duplicate-heavy rejection feedback in one construction round, the Critic summarizes the failure pattern and produces query-diversification guidance for the next round.
    In this logged case, deduplication failures decrease from 278 to 148 after the update.
    }
    \label{fig:critic_case}
\end{figure}

To stabilize iterative refinement, \method maintains an experience library that stores high-value semantic feedback and effective query patterns from recent rounds.
The library is updated by retaining useful feedback entries and merging near-duplicate observations, so that later retrieval planning can reuse relevant experience without accumulating redundant memory.

Conditioned on semantic feedback, the Critic guides query diversification and generation-prompt refinement through language-based policy updates.
The experience library mainly supports retrieval-side reuse, while prompt refinement for generated samples is driven by recent verification feedback, such as OCR failures, semantic mismatches, or visual-quality issues.
In practice, this LLM-based feedback summarizer converts round-level statistics and rejection patterns into actionable retrieval guidance and prompt-level revision cues.
Figure~\ref{fig:critic_case} illustrates the retrieval-side update process with a duplicate-heavy construction round, where rejection statistics are converted into semantic guidance for next-round query diversification.

\subsection{Generator: Coverage Gaps to Synthetic Completion}
\label{sec:generator_completion}

Retrieval alone is often insufficient to fully cover the target distribution, especially for rare topics, complex layouts, long-tail text patterns, or structured visual scenarios.
To address this limitation, \method introduces a Generator that transforms \emph{coverage gaps} into \emph{targeted synthetic completion}.

After verification, the system measures the coverage of each topic--subtopic pair in the current pass set:
\[
G_t(u,v)
=
\left|
\left\{
i \in \mathcal{I}_{\mathrm{pass}}^t
:
\mathrm{topic}(i)=u,\ 
\mathrm{subtopic}(i)=v
\right\}
\right|,
\]
where $(u,v)$ denotes a topic and subtopic.
Regions with coverage significantly below the current distributional average, or below a predefined minimum coverage count, are flagged as underrepresented.

The Generator then synthesizes candidate samples for these under-covered regions.
Specifically, generation prompts are initialized from the target topic--subtopic pair and the required completion count, so that the generated candidates are directed toward regions with insufficient coverage.
This design complements retrieval-based collection by actively supplying samples for long-tail or sparsely covered semantic regions.

Generated samples are passed through the same fixed verification pipeline as retrieved candidates before being admitted to the dataset.
If a class of generated samples repeatedly fails in OCR quality, semantic alignment, or visual quality, those failures are recorded as part of the round-level rejection statistics.
The Critic summarizes these generated-sample failures into semantic feedback, which is then used to refine subsequent generation prompts.
Through repeated iterations, \method not only selects better samples from available sources, but also actively fills structural gaps in the target data distribution.

\section{Experiments}
\label{sec:experiments}

\subsection{Experimental Setup}
\label{sec:exp_setup}

\paragraph{Task and objective.}
We evaluate \method in the setting of text-rich image data construction.
The goal of our experiments is not to introduce a new downstream generator, but to examine whether a self-evolving data construction process can produce more useful training data than fixed-dataset baselines under matched data budgets.
We organize the experiments around four questions: (i) whether \method improves downstream text-rich image generation across different downstream models and benchmarks, (ii) whether the Critic and Generator are necessary components of the closed-loop construction process, (iii) whether the advantage of \method persists under increasing data budgets, and (iv) how the framework changes construction-time data quality and failure patterns.

\paragraph{Training data and baselines.}
We compare \method with two strong public text-rich image data sources, MARIO-10M~\citep{textdiffuser} and AnyWord-3M~\citep{anytext}.
For fair comparison, all methods are evaluated under matched data budgets.
Unless otherwise specified, the main comparison uses the same number of training samples for each data source.
For module ablation, we construct two variants of \method: \textbf{Ours w/o Critic}, which removes semantic feedback and policy revision, and \textbf{Ours w/o Generator}, which removes targeted synthetic completion and relies only on retrieval and verification.

\begin{table}[t]
\centering
\scriptsize
\setlength{\tabcolsep}{2.4pt}
\caption{
Main results on TextScenesHQ and LongTextBench under the 0.75M matched data budget.
Numbers in parentheses denote absolute F1 gains over the strongest baseline in each setting.
FID is unavailable for LongTextBench due to the absence of reference images and is marked as ``--''.
}
\label{tab:main_results}
\resizebox{0.985\linewidth}{!}{
\begin{tabular}{L{2.15cm}L{1.50cm}L{1.08cm}L{0.88cm}L{0.98cm}L{1.13cm}L{1.13cm}L{1.35cm}}
\toprule
\textbf{Benchmark} & \textbf{Method} & \textbf{CLIP} $\uparrow$ & \textbf{FID} $\downarrow$ & \textbf{Acc.} $\uparrow$ & \textbf{Prec.} $\uparrow$ & \textbf{Recall} $\uparrow$ & \textbf{F1} $\uparrow$ \\
\midrule
\multicolumn{8}{c}{\textbf{PixArt-$\alpha$}} \\
\midrule
\multirow{3}{*}{TextScenesHQ}
& AnyWord & 0.264 & 71 & 0.34 & 4.76 & 0.34 & 0.63 \\
& MARIO   & \textbf{0.274} & 76 & 2.61 & \textbf{18.07} & 2.61 & 4.56 \\
& \cellcolor{lightblue}Ours
& \cellcolor{lightblue}\textbf{0.274}
& \cellcolor{lightblue}\textbf{68}
& \cellcolor{lightblue}\textbf{6.01}
& \cellcolor{lightblue}14.21
& \cellcolor{lightblue}\textbf{6.01}
& \cellcolor{lightblue}\textbf{8.45} \gain{3.89} \\
\cmidrule(lr){1-8}
\multirow{3}{*}{LongTextBench}
& AnyWord & \textbf{0.209} & -- & 0.30 & 2.77 & 0.30 & 0.54 \\
& MARIO   & 0.194 & -- & 4.48 & \textbf{13.35} & 4.48 & 6.71 \\
& \cellcolor{lightblue}Ours
& \cellcolor{lightblue}0.196
& \cellcolor{lightblue}--
& \cellcolor{lightblue}\textbf{6.92}
& \cellcolor{lightblue}13.22
& \cellcolor{lightblue}\textbf{6.92}
& \cellcolor{lightblue}\textbf{9.08} \gain{2.37} \\
\midrule
\multicolumn{8}{c}{\textbf{Show-o2}} \\
\midrule
\multirow{3}{*}{TextScenesHQ}
& AnyWord & 0.263 & 81 & 0.08 & 2.90 & 0.08 & 0.16 \\
& MARIO   & 0.262 & 68 & 0.10 & 2.00 & 0.10 & 0.19 \\
& \cellcolor{lightblue}Ours
& \cellcolor{lightblue}\textbf{0.271}
& \cellcolor{lightblue}\textbf{64}
& \cellcolor{lightblue}\textbf{0.24}
& \cellcolor{lightblue}\textbf{3.20}
& \cellcolor{lightblue}\textbf{0.24}
& \cellcolor{lightblue}\textbf{0.45} \gain{0.26} \\
\cmidrule(lr){1-8}
\multirow{3}{*}{LongTextBench}
& AnyWord & 0.206 & -- & 0.11 & 4.00 & 0.11 & 0.21 \\
& MARIO   & 0.210 & -- & 0.14 & 3.60 & 0.14 & 0.27 \\
& \cellcolor{lightblue}Ours
& \cellcolor{lightblue}\textbf{0.211}
& \cellcolor{lightblue}--
& \cellcolor{lightblue}\textbf{0.23}
& \cellcolor{lightblue}\textbf{4.50}
& \cellcolor{lightblue}\textbf{0.23}
& \cellcolor{lightblue}\textbf{0.44} \gain{0.17} \\
\bottomrule
\end{tabular}
}
\end{table}

\paragraph{Construction implementation.}
In our implementation, query generation and coverage planning are handled by Mistral-7B.
The ExperienceLibrarian, SemanticFeedback, and PromptPlanner agents use Qwen3.5-4B.
Targeted synthetic image generation is performed with Qwen-Image, with prompts initialized from the target topic--subtopic pair and required completion count and refined using Critic feedback from generated-sample verification.
Qwen3-VL is used only for optional caption/annotation of accepted samples, not for pass/fail verification decisions.
Generated and retrieved samples are evaluated with the same fixed Verifier configuration.
All variants use the same verification protocol and matched data budget.
OCR quality is measured with PaddleOCR~\citep{paddleocr}, while semantic consistency is evaluated with CLIP ViT-B/32~\citep{clip}.
These signals are summarized into round-level feedback for later policy revision.

\paragraph{Downstream models.}
To examine whether the constructed data generalizes beyond a single generator, we fine-tune two downstream text-rich image generation models, PixArt-$\alpha$~\citep{pixartalpha} and Show-o2~\citep{showo2}.
All models are trained under matched data budgets and evaluated with the same benchmark prompts.
Unless otherwise specified, images are resized to $512 \times 512$.
Detailed downstream training configurations are provided in Appendix~\ref{app:downstream_training}.

\paragraph{Evaluation benchmarks and metrics.}
We evaluate the fine-tuned models on TextScenesHQ and LongTextBench.
TextScenesHQ focuses on scene-level text rendering quality, while LongTextBench emphasizes longer textual content and more complex text-layout interactions.
We report CLIP Score~\citep{clip} for image-text semantic alignment and FID~\citep{fid} for visual quality when reference images are available.
For text rendering quality, we report OCR accuracy, precision, recall, and F1 score.
Following our evaluation protocol, OCR matching is computed with a fuzzy matching threshold of 70.
Since OCR-F1 balances precision and recall, we treat it as the primary metric.

\subsection{Main Results}
\label{sec:main_results}

Table~\ref{tab:main_results} reports the main comparison under the 0.75M matched data budget.
On PixArt-$\alpha$, \method achieves the \textbf{best OCR-F1 on both benchmarks}, improving over the strongest baseline from \textbf{4.56 to 8.45} on TextScenesHQ and from \textbf{6.71 to 9.08} on LongTextBench.
On Show-o2, \method also obtains the \textbf{strongest OCR-oriented performance}, improving F1 from \textbf{0.19 to 0.45} on TextScenesHQ and from \textbf{0.27 to 0.44} on LongTextBench.
These results show that the benefit of \method is \textbf{not tied to a single downstream generator}.

The detailed metrics further indicate that \method mainly improves \emph{text coverage} and \emph{recognizability}.
Across all four settings, \method achieves the \textbf{highest OCR accuracy, recall, and F1}.
On PixArt-$\alpha$, MARIO obtains slightly higher OCR precision, but its lower recall and F1 suggest that it produces fewer correctly matched text instances overall.
In contrast, \method improves recall substantially while maintaining competitive precision, leading to stronger balanced OCR performance.

\begin{figure}[t]
    \centering
    \includegraphics[width=0.85\linewidth]{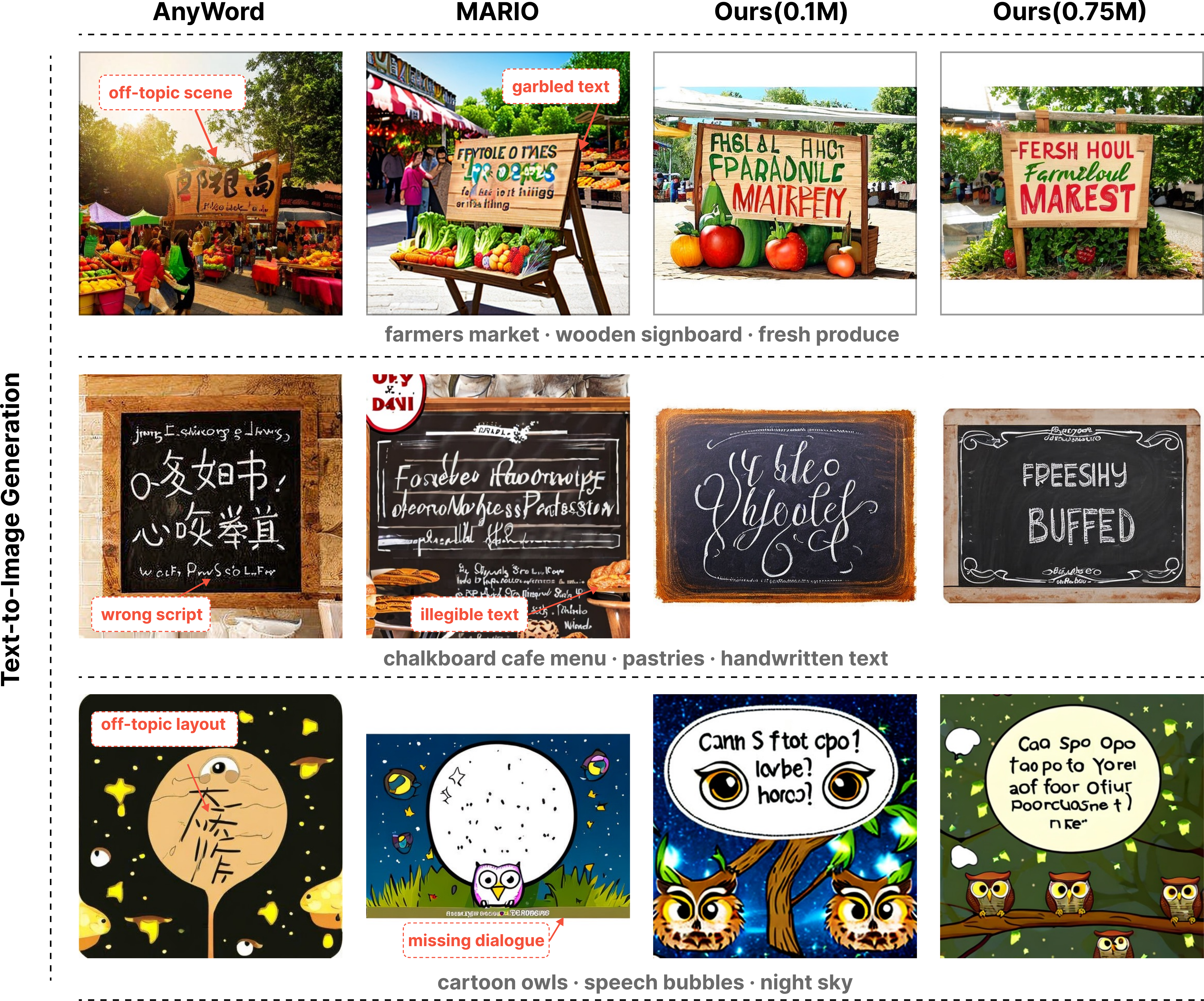}
    \caption{
    \textbf{Qualitative comparison of downstream generations.}
    Representative examples show that \method alleviates common failure modes, including off-topic generations, missing text regions, and severe text distortion. Text rendering remains challenging in long or stylized cases.
    }
    \label{fig:showcase}
\end{figure}

CLIP Score does not always follow the same trend as OCR-F1, which is expected because global semantic alignment and text legibility measure different aspects of text-rich image generation.
For example, AnyWord obtains the highest CLIP Score on PixArt-$\alpha$ LongTextBench, while \method still achieves the best OCR-F1.
When FID is available, \method also obtains the \textbf{lowest FID} on TextScenesHQ for both downstream models.
Qualitative examples in Figure~\ref{fig:showcase} further show that \method reduces off-topic generations, missing text regions, and severe text distortion.

\subsection{Ablation and Process Analysis}
\label{sec:module_ablation}

We conduct module ablations to examine whether the Critic and Generator are necessary for effective data construction.
The Critic converts round-level feedback and failure patterns into semantic policy updates, while the Generator performs targeted synthetic completion for under-covered regions.
All variants use the same Verifier, the same downstream training protocol, and the same data budget, so the comparison isolates the effect of the removed construction component.

\begin{table}[htbp]
\centering
\scriptsize
\setlength{\tabcolsep}{3.0pt}
\caption{
\textbf{Module ablation on PixArt-$\alpha$ at the 0.1M scale.}
}
\label{tab:ablation}
\resizebox{0.985\linewidth}{!}{
\begin{tabular}{L{2.15cm}L{1.50cm}L{1.08cm}L{0.88cm}L{0.98cm}L{1.13cm}L{1.13cm}L{0.88cm}}
\toprule
\textbf{Benchmark} & \textbf{Method} & \textbf{CLIP} $\uparrow$ & \textbf{FID} $\downarrow$ & \textbf{Acc.} $\uparrow$ & \textbf{Prec.} $\uparrow$ & \textbf{Recall} $\uparrow$ & \textbf{F1} $\uparrow$ \\
\midrule
\multirow{3}{*}{TextScenesHQ}
& \cellcolor{lightblue}Ours
& \cellcolor{lightblue}\textbf{0.278}
& \cellcolor{lightblue}\textbf{59}
& \cellcolor{lightblue}\textbf{1.03}
& \cellcolor{lightblue}\textbf{6.41}
& \cellcolor{lightblue}\textbf{1.03}
& \cellcolor{lightblue}\textbf{1.78} \\
& w/o Critic    & 0.270 & 63 & 0.56 & 5.31 & 0.56 & 1.01 \\
& w/o Generator & 0.273 & 61 & 0.79 & 5.86 & 0.79 & 1.40 \\
\cmidrule(lr){1-8}
\multirow{3}{*}{LongTextBench}
& \cellcolor{lightblue}Ours
& \cellcolor{lightblue}0.192
& \cellcolor{lightblue}--
& \cellcolor{lightblue}\textbf{1.51}
& \cellcolor{lightblue}\textbf{3.81}
& \cellcolor{lightblue}\textbf{1.51}
& \cellcolor{lightblue}\textbf{2.16} \\
& w/o Critic    & \textbf{0.203} & -- & 0.58 & 2.07 & 0.58 & 0.90 \\
& w/o Generator & 0.192 & -- & 1.02 & 2.07 & 1.02 & 1.37 \\
\bottomrule
\end{tabular}
}
\end{table}

As shown in Table~\ref{tab:ablation}, removing either module degrades downstream OCR-F1, but the two ablated variants fail in different ways.
Removing the Critic causes the \textbf{larger drop}, reducing F1 from \textbf{1.78 to 1.01} on TextScenesHQ and from \textbf{2.16 to 0.90} on LongTextBench.
This indicates that \textbf{verification signals alone are insufficient}: without a module that turns rejection causes into policy-level feedback, later construction rounds cannot effectively avoid recurring failure modes.
Removing the Generator also weakens performance, reducing F1 to \textbf{1.40} on TextScenesHQ and \textbf{1.37} on LongTextBench.
The smaller but consistent drop suggests that targeted synthetic completion mainly improves \textbf{coverage of under-represented regions}, while the Critic is more directly responsible for making the construction policy \textbf{adaptive}.

\FloatBarrier

\vspace{0.6em}
\noindent
\begin{minipage}[t]{0.49\linewidth}
\centering
\captionof{table}{
\textbf{Construction-time statistics at the 0.1M scale.}
}
\label{tab:construction_stats}
\vspace{-0.4em}
\scriptsize
\setlength{\tabcolsep}{3.5pt}
\resizebox{\linewidth}{!}{
\begin{tabular}{L{2.05cm}C{1.25cm}C{1.45cm}C{1.45cm}}
\toprule
\textbf{Variant} & \textbf{Pass} $\uparrow$ & \textbf{OCR Conf.} $\uparrow$ & \textbf{Coverage} $\uparrow$ \\
\midrule
w/o Critic    & 0.532 & 0.861 & 0.935 \\
w/o Generator & 0.608 & 0.914 & 0.818 \\
\cellcolor{lightblue}Ours
& \cellcolor{lightblue}\textbf{0.671}
& \cellcolor{lightblue}\textbf{0.938}
& \cellcolor{lightblue}\textbf{0.961} \\
\bottomrule
\end{tabular}
}
\end{minipage}
\hfill
\begin{minipage}[t]{0.49\linewidth} 
\centering 
\captionof{table}{ 
\textbf{Sensitivity to the Critic backbone on TextScenesHQ at the 10k scale.} 
} 
\label{tab:critic_sensitivity} 
\vspace{-0.4em} 
\scriptsize \setlength{\tabcolsep}{4.0pt} 
\resizebox{\linewidth}{!}{ 
\begin{tabular}{L{2.45cm}C{1.45cm}C{1.45cm}} 
\toprule 
\textbf{Critic Backbone} & \textbf{OCR Acc.} $\uparrow$ & \textbf{OCR-F1} $\uparrow$ \\ 
\midrule 
Qwen3.5-4B & 0.41 & 0.75 \\ 
Qwen3.5-35B & \textbf{0.66} & \textbf{1.14} \\ 
\bottomrule 
\end{tabular} } 
\end{minipage}

\vspace{0.6em}
The construction-time statistics in Table~\ref{tab:construction_stats} explain the downstream ablation results.
The full framework achieves the highest pass rate, OCR confidence, and topic coverage, showing that its advantage already emerges during data construction.
Without the Critic, coverage remains relatively high, but both pass rate and OCR confidence drop substantially, indicating that broad coverage alone does not ensure high-quality text-rich supervision.
Without the Generator, OCR confidence is higher than the no-Critic variant, but coverage becomes much lower, suggesting that retrieval and verification can select cleaner samples but are less effective at filling long-tail or under-covered regions.
These results support the complementary roles of the two modules: the Critic improves the direction of policy revision, while the Generator expands data coverage.
Table~\ref{tab:critic_sensitivity} further reports a lightweight Critic-backbone sensitivity analysis on TextScenesHQ at the 10k scale.
We only replace the Critic model while keeping the other agents, verification settings, and training budget fixed.
Using Qwen3.5-35B as the Critic improves OCR accuracy from 0.41 to 0.66 and OCR-F1 from 0.75 to 1.14.
This suggests that a stronger Critic can further improve feedback-driven data construction, while the 4B Critic remains a lightweight and effective default choice.

\vspace{0.4em}

\begin{wrapfigure}{r}{0.43\linewidth}
\vspace{-1.0em}
\centering
\includegraphics[width=0.98\linewidth]{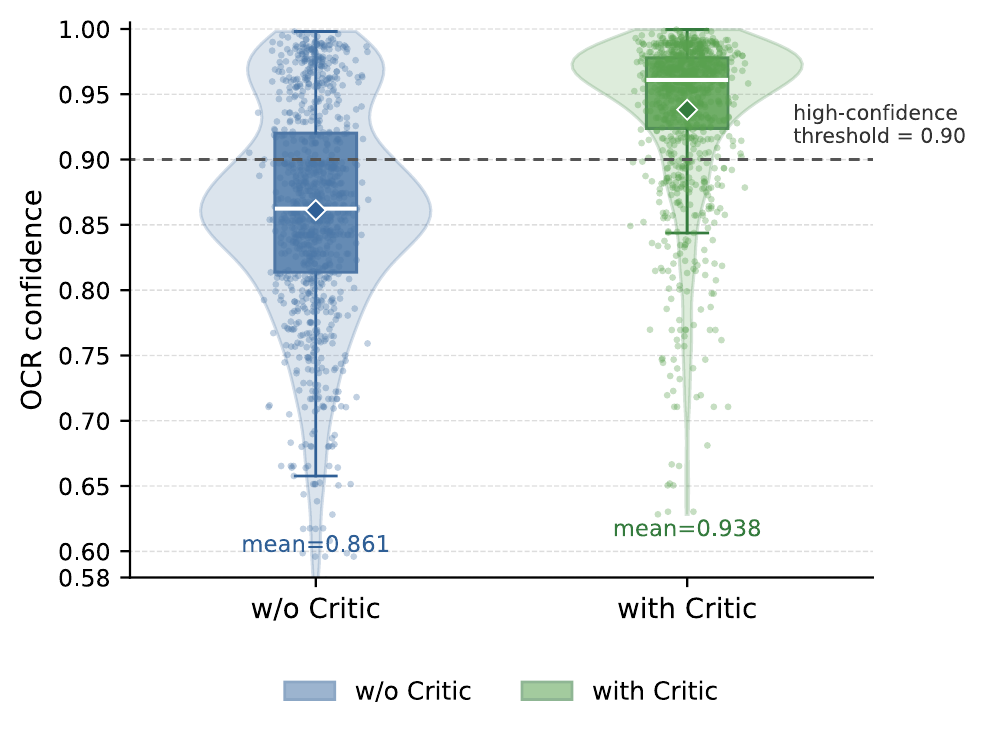}
\vspace{-0.7em}
\caption{
\textbf{Construction-time OCR confidence with and without the Critic.}
}
\label{fig:ocr_confidence}
\vspace{-1.0em}
\end{wrapfigure}

Beyond aggregate statistics, Figure~\ref{fig:ocr_confidence} compares the distribution of OCR confidence for accepted samples with and without the Critic.
Enabling the Critic increases mean OCR confidence from \textbf{0.861 to 0.938}, and the proportion of high-confidence samples above 0.90 rises from \textbf{29.1\% to 81.1\%}.
This distributional shift provides process-level evidence that the Critic does not merely improve final model scores indirectly.
Instead, it changes the accepted data itself by making recurring recognition failures actionable in later construction rounds.
Such improvement is especially relevant for text-rich image generation, where a small number of low-confidence text regions can substantially reduce OCR-oriented metrics even when the image remains semantically plausible.

\vspace{0.4em}

\begin{wrapfigure}{l}{0.40\linewidth}
\vspace{-0.8em}
\centering
\includegraphics[width=0.98\linewidth]{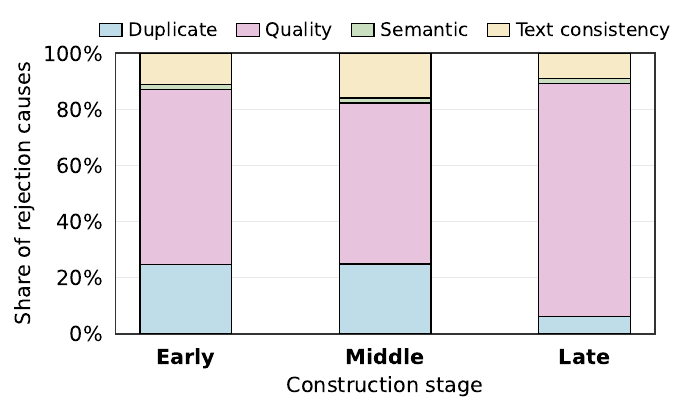}
\vspace{-0.7em}
\caption{
\textbf{Stage-level rejection composition over construction rounds.}
}
\label{fig:rejection_composition}
\vspace{-1.0em}
\end{wrapfigure}

Figure~\ref{fig:rejection_composition} further analyzes how rejection causes evolve over construction stages.
We aggregate sampled construction rounds into Early, Middle, and Late stages and normalize each stacked bar to 100\%, so the figure focuses on the relative composition of failure modes rather than the absolute number of rejected samples.
This diagnostic complements the OCR confidence analysis by characterizing the remaining failures within the evolving construction process.
The composition indicates that different rejection causes dominate at different stages, suggesting that construction-time feedback exposes \textbf{structured failure patterns} rather than random filtering noise.
As easier failures are reduced, the remaining rejected samples become concentrated in harder residual issues such as visual quality and text consistency.
This supports the central premise of \method: rejected samples contain \textbf{reusable information} for revising the construction policy instead of being discarded as uninformative byproducts.

\FloatBarrier

\subsection{Scaling Trend}
\label{sec:scaling_trend}

To study how different data construction strategies behave under increasing data budgets, we conduct a scaling trend analysis over multiple data scales.
For each scale, we compare \method with matched-scale subsets from MARIO-10M and AnyWord-3M.

\begin{table}[!htbp]
\centering
\footnotesize
\setlength{\tabcolsep}{8pt}
\renewcommand{\arraystretch}{1.05}
\caption{
\textbf{Scaling trend of F1 under matched data budgets.}
}
\label{tab:scaling}
\begin{tabular}{L{2.25cm}L{1.65cm}C{1.10cm}C{1.10cm}C{1.10cm}C{1.10cm}}
\toprule
\textbf{Benchmark} & \textbf{Method} & \textbf{0.1M} & \textbf{0.25M} & \textbf{0.5M} & \textbf{0.75M} \\
\midrule
\multicolumn{6}{c}{\textbf{PixArt-$\alpha$}} \\
\midrule
\multirow{3}{*}{TextScenesHQ}
& AnyWord & 0.42 & 0.47 & 0.53 & 0.63 \\
& MARIO   & 0.56 & 1.24 & 3.55 & 4.56 \\
& \cellcolor{lightblue}Ours
& \cellcolor{lightblue}\textbf{1.78}
& \cellcolor{lightblue}\textbf{4.17}
& \cellcolor{lightblue}\textbf{6.67}
& \cellcolor{lightblue}\textbf{8.45} \\
\cmidrule(lr){1-6}
\multirow{3}{*}{LongTextBench}
& AnyWord & 0.28 & 0.21 & 0.47 & 0.54 \\
& MARIO   & 0.79 & 1.08 & 5.76 & 6.71 \\
& \cellcolor{lightblue}Ours
& \cellcolor{lightblue}\textbf{2.16}
& \cellcolor{lightblue}\textbf{4.04}
& \cellcolor{lightblue}\textbf{8.35}
& \cellcolor{lightblue}\textbf{9.08} \\
\midrule
\multicolumn{6}{c}{\textbf{Show-o2}} \\
\midrule
\multirow{3}{*}{TextScenesHQ}
& AnyWord & 0.08 & 0.09 & 0.12 & 0.16 \\
& MARIO   & 0.11 & 0.09 & 0.16 & 0.19 \\
& \cellcolor{lightblue}Ours
& \cellcolor{lightblue}\textbf{0.19}
& \cellcolor{lightblue}\textbf{0.30}
& \cellcolor{lightblue}\textbf{0.39}
& \cellcolor{lightblue}\textbf{0.45} \\
\cmidrule(lr){1-6}
\multirow{3}{*}{LongTextBench}
& AnyWord & 0.05 & 0.14 & 0.18 & 0.21 \\
& MARIO   & 0.04 & 0.13 & 0.23 & 0.27 \\
& \cellcolor{lightblue}Ours
& \cellcolor{lightblue}\textbf{0.18}
& \cellcolor{lightblue}\textbf{0.30}
& \cellcolor{lightblue}\textbf{0.36}
& \cellcolor{lightblue}\textbf{0.44} \\
\bottomrule
\end{tabular}
\end{table}

As shown in Table~\ref{tab:scaling}, \method \textbf{consistently outperforms both fixed baselines} across the completed data scales.
On PixArt-$\alpha$, \method improves from 1.78 to 4.17, 6.67, and 8.45 on TextScenesHQ, and from 2.16 to 4.04, 8.35, and 9.08 on LongTextBench as the budget increases from 0.1M to 0.75M.
The same trend is observed on Show-o2, although the absolute values are lower.
We interpret these results as a scaling trend rather than a strict scaling law, since formal scaling-law fitting would require more data points and repeated runs.

\subsection{Semantic Diversity Analysis}
\label{sec:diversity_analysis}

One potential concern is that the improvement of \method may come from concentrating on a narrow set of high-quality text-rich image scenarios.
To examine this, we further analyze whether the constructed data covers diverse text-rich image categories.
Since MARIO and AnyWord do not provide unified fine-grained category annotations, we define a fixed text-rich image taxonomy and apply the same zero-shot caption classifier~\citep{laurer2023building} to MARIO, AnyWord, and \method.
We focus on two coverage-oriented statistics: category coverage and tail coverage.
More details of the taxonomy, classifier, and metric definitions are provided in Appendix~\ref{app:diversity_details}.

\begin{table}[!htbp]
\centering
\footnotesize
\setlength{\tabcolsep}{8pt}
\renewcommand{\arraystretch}{1.05}
\caption{
\textbf{Coverage-oriented semantic diversity statistics at the 0.5M scale.}
}
\label{tab:semantic_diversity}
\begin{tabular}{L{4.1cm}C{1.7cm}C{1.7cm}C{1.7cm}}
\toprule
\textbf{Metric} & \textbf{MARIO} & \textbf{AnyWord} & \textbf{Ours} \\
\midrule
Category Coverage $\uparrow$
& 90.91\%
& 78.79\%
& \cellcolor{lightblue}\textbf{96.97\%} \\
Tail Coverage $\uparrow$
& 1.76\%
& 0.61\%
& \cellcolor{lightblue}\textbf{2.45\%} \\
\bottomrule
\end{tabular}
\end{table}

As shown in Table~\ref{tab:semantic_diversity}, \method achieves the highest category coverage and tail coverage among the compared datasets.
These results suggest that the feedback-driven construction process helps recover a broader set of text-rich image scenarios and improves the inclusion of long-tail categories under the same data budget.

% \FloatBarrier
\section{Conclusion and Limitations}

We presented \method, a self-evolving multi-agent framework for text-rich image data construction.
Rather than treating dataset construction as a static crawl--filter--freeze pipeline, \method turns construction-time failures into semantic feedback for later rounds. Through the Retriever, Verifier, Critic, and Generator, the framework forms a closed loop that uses fixed verification feedback to refine retrieval queries, generation prompts, and experience memory during data construction.

Experiments on text-rich image generation show that this feedback-driven construction process produces more useful training data than fixed-dataset baselines under matched data budgets.
The ablation results further indicate that the Critic and Generator play different but complementary roles.
The Critic makes rejection feedback actionable for policy revision, while the Generator improves coverage by completing under-represented regions.
These results support the view that data construction can be treated as an adaptive process rather than a fixed preprocessing step.

\textbf{Limitations.}
Our scaling results are better interpreted as a trend rather than a strict scaling law, since formal scaling-law analysis would require more data scales and repeated runs.
The framework also depends on the reliability of the Verifier, so noisy optical character recognition, semantic scoring, or image-quality estimation may affect later policy updates.
In addition, this paper focuses on text-rich image generation, while extending the same feedback-driven construction paradigm to broader multimodal domains remains an important direction for future work.

%%%%%%%%%%%%%%%%%%%%%%%%%%%%%%%%%%%%%%%%%%%%%%%%%%%%%%%%%%%%

\appendix

\bibliographystyle{plainnat}
\bibliography{main}

%%%%%%%%%%%%%%%%%%%%%%%%%%%%%%%%%%%%%%%%%%%%%%%%%%%%%%%%%%%%

\appendix
\clearpage
\appendix

\newtcolorbox{promptbox}[1]{
  enhanced,
  breakable,
  colback=white,
  colframe=black!45,
  boxrule=0.4pt,
  arc=4pt,
  left=6pt,
  right=6pt,
  top=6pt,
  bottom=6pt,
  title=#1,
  colbacktitle=black,
  coltitle=white,
  fonttitle=\bfseries,
  toptitle=2pt,
  bottomtitle=2pt
}

%%%%%%%%%%%%%%%%%%%%%%%%%%%%%%%%%%%%%%%%%%%%%%%%%%%%%%%%%%%%
\section{Implementation Details}
\label{app:implementation}

\subsection{Verifier Configuration}
OCR is first applied to extract text signals from each candidate image.
The filtering pipeline then applies perceptual-hash deduplication, OCR/image-quality assessment, CLIP-based semantic relevance checking, and OCR-text consistency checking.
Semantic relevance is measured with CLIP ViT-B/32, while OCR-text consistency is checked with Sentence-Transformer (all-MiniLM-L6-v2).
The resulting rejection statistics are aggregated into round-level feedback and used by the Critic for subsequent policy updates.
The Verifier follows a fixed evaluation pipeline and does not use Qwen3-VL for pass/fail verification decisions.

\subsection{Downstream Training Details}
\label{app:downstream_training}

We fine-tune PixArt-$\alpha$ and Show-o2 under matched data budgets for all compared
data sources. Unless otherwise specified, input images are resized to $512 \times 512$.
Unset options follow the default settings of the corresponding training code.

\paragraph{PixArt-$\alpha$.}
For PixArt-$\alpha$, we use the \texttt{PixArt\_XL\_2} architecture initialized from
the official PixArt-XL-2 512$\times$512 checkpoint, together with the SD VAE
\texttt{sd-vae-ft-ema}. We disable window attention by setting the window size to
0 and do not use relative positional encoding. Attention is computed in FP32.
We train with a batch size of 32, gradient accumulation step of 1, and 10 data
loading workers. Gradient checkpointing is enabled, and gradient clipping is set
to 0.01. The optimizer is AdamW with learning rate $1\times10^{-4}$, weight decay
$3\times10^{-2}$, and $\epsilon=1\times10^{-10}$. We use linear warmup for 1000
steps. The logging interval is 20 steps, and model checkpoints are saved every
2000 steps.

\paragraph{Show-o2.}
For Show-o2, we use AdamW with separate learning rates for different modules:
$2\times10^{-6}$ for the visual encoder, $1\times10^{-5}$ for the projection
module, and $1\times10^{-5}$ for the Show-o backbone. We set $\beta_1=0.9$,
$\beta_2=0.999$, weight decay to 0, and $\epsilon=1\times10^{-8}$. The learning
rate follows a cosine schedule with a warmup ratio of 0.03. Training uses BF16
mixed precision with TF32 enabled, gradient accumulation step of 16, and batch
size 1 for both text-to-image and multimodal-understanding batches. The maximum
gradient norm is set to 1.0, the conditional dropout probability is 0.1, and the
training seed is 1008. The model is trained in the tuning stage with flow loss
coefficient 1.0 and next-token prediction coefficient 0.0.

For the flow-matching transport configuration, we use a linear path type with
velocity prediction and log-normal SNR weighting. During sampling, we use Euler
sampling with guidance scale 5.0 and 50 inference steps. The absolute and relative
tolerances are set to $10^{-6}$ and $10^{-3}$, respectively. Time shifting is
enabled with a shifting factor of 3.0.

\subsection{Semantic Diversity Evaluation Details}
\label{app:diversity_details}

We evaluate semantic diversity using the captions associated with each image.
All compared datasets are processed with the same zero-shot caption classifier,
\texttt{MoritzLaurer/deberta-v3-large-zeroshot-v2.0}~\citep{laurer2023building}.
For each caption, the classifier predicts one label from a fixed taxonomy of
text-rich image scenarios. If the maximum classification confidence is below
0.15, the sample is assigned to an additional \textit{other} category.
The \textit{other} category is used only as a fallback label and is excluded
from the reported diversity statistics.

We report two coverage-oriented statistics in the main paper.
Category coverage is the percentage of predefined categories whose frequency is
higher than 0.1\%.
Tail coverage is the cumulative frequency of the ten least frequent predefined
categories.
Both statistics are computed over the 33 predefined candidate categories after
excluding samples assigned to the \textit{other} category.
All datasets are evaluated at the 0.5M scale using the same taxonomy, classifier,
confidence threshold, and metric definitions.

\begin{table}[!htbp]
\centering
\small
\setlength{\tabcolsep}{6pt}
\renewcommand{\arraystretch}{1.15}
\caption{
\textbf{Grouped taxonomy used for semantic diversity evaluation.}
}
\label{tab:semantic_taxonomy}
\begin{tabular}{L{3.1cm}L{9.1cm}}
\toprule
\textbf{Group} & \textbf{Candidate labels} \\
\midrule
\textbf{Documents and records}
& certificate, record, form, manual, statement, resume, contract, policy,
transcript, ballot, patent, deed, tax form, specification sheet \\
\midrule
\textbf{Commercial and transactional text}
& waybill, invoice, ticket, card, label, product packaging, menu, receipt \\
\midrule
\textbf{Printed and public text}
& signage, poster, book, research paper, slide, handwritten note \\
\midrule
\textbf{Digital and social media}
& UI screenshot, webpage, social media post, meme, comic \\
\bottomrule
\end{tabular}
\end{table}

\FloatBarrier

% \clearpage

%%%%%%%%%%%%%%%%%%%%%%%%%%%%%%%%%%%%%%%%%%%%%%%%%%%%%%%%%%%%
\section{Prompts}
\label{app:prompts}

This appendix presents the prompt templates used in DataEvolver.

\subsection{Retrieval and Generation Prompts}
\label{app:retrieval_generation_prompts}

\begin{promptbox}{Prompt for Coverage Planning}
You are an image dataset quantity balancing analyst. Your task is to plan the number of images needed for each theme-subtheme based on the total target and current data distribution.\\

Strictly follow the requirements and return only the JSON array (string list), without any explanations:\\

1) Only output subtopics that need additional images; for subtopics that have already reached or exceeded a reasonable number, return 0 or omit them.\\
2) Do not output negative numbers or decimals.\\
3) Output should be a string array, with each element formatted as 'subtopic name: need X more images'.\\
4) Output format: JSON array, without any explanations, e.g., ['TopicA SubtopicA: 50', 'TopicA SubtopicB: 30', 'SubtopicC: 20'].
\end{promptbox}

\begin{promptbox}{Prompt for Retriever}
You are a search strategy generator. Given a subtopic, construct up to 3 Bing image queries and return them as a JSON array.\\

Strong constraints:\\

1) Each query must contain keywords that retrieve images with readable text: label, packaging label, nutrition facts, ingredients, barcode, instruction, manual, leaflet, receipt, certificate, signage, timetable, menu, UI screenshot.\\
2) It is recommended to add reliable sources such as site:wikimedia.org, site:flickr.com, or site:openfoodfacts.org.\\
3) Emphasize clarity and richness with terms such as high resolution, 4k, detailed, rich texture, HD, and with readable text.\\
4) Avoid simple, minimalist, or plain-background images.\\
5) Return only the JSON array without any explanation.\\

Subtheme: \{subtopic\}\\
Generate up to 3 Bing search queries highly relevant to '\{subtopic\}'.
\end{promptbox}

\begin{promptbox}{Prompt for Query Diversification}
You are a diversified search strategy generator. Given a subtopic, generate up to 5 distinct search queries and return them as a JSON array.\\

Strong constraints:\\

1) Each query must explicitly include a text-related description, such as readable text, packaging label, nutrition facts, ingredients list, back-of-package text, document text, instruction sheet, receipt text, or signage text.\\
2) Emphasize no watermark, clean layout, high contrast, high resolution, and rich visual details.\\
3) Avoid simple, minimalist, or plain-background images.\\
4) Return only the JSON array without any explanation.\\

Generate up to 5 diverse search queries for '\{subtopic\}'.
\end{promptbox}

\begin{promptbox}{Prompt for Generator Prompt Planning}
You are an image generation prompt engineer. Your goal is to generate high-quality image generation prompts for a specified theme-subtheme, in the required number.\\

Strictly follow these requirements and return only the JSON array (string list), without any explanations:\\

1) The prompts should combine the theme and subtheme, highlighting key elements of the text scene, including keywords, text content, visual scene, and quality features.\\
2) Each prompt must ensure that the generated image includes visible, OCR-readable text, such as billboard, packaging, menu, poster, book page, etc.\\
3) Each prompt should generate a concise statement, clear and direct, without additional explanation.\\
4) Output format: ['prompt1', 'prompt2', 'prompt3', ...].\\

Runtime input:\\

Based on the following information, generate high-quality English image generation prompts. Only return a JSON array, list of strings.\\

Theme: \{topic\}\\
Subtheme: \{subtopic\}\\
Number of prompts needed: \{need\}\\

Generation Requirements:\\
1) The prompt must focus on the theme, listing key textual elements that must appear in the image.\\
2) Text must be clear, high contrast, watermark-free.\\
3) Layout natural and neat; avoid clutter.\\
4) For complex scenes request simpler layout if needed.\\
5) Output only the JSON array of prompt strings.
\end{promptbox}

\subsection{Critic and Experience-Library Prompts}
\label{app:critic_memory_prompts}

\begin{promptbox}{Prompt for Semantic Feedback}
INPUT DATA\\

1. Current Round Rejections:\\
\{current\_rejection\_counts\}\\

2. Previous Round Rejections:\\
\{prev\_rejection\_counts\}\\

3. Keyword Performance, accepted vs. rejected:\\
\{keyword\_analysis\}\\

TASK\\

You are a critical analyst. Produce a highly abstract, concise semantic feedback summary of 4--6 sentences.\\

1. Keyword Optimization:\\
- Identify keywords from keyword\_analysis that appear mostly in rejected items, and suggest weakening or removing them.\\
- Identify keywords that appear in accepted items, and suggest strengthening them.\\

OUTPUT FORMAT\\

1. Output exactly one continuous natural-language paragraph, 4--6 sentences.\\
2. No bullet points, no headings, no Markdown formatting, no JSON.\\
3. Required style: ``Blurry sample ratio increased (Prev: 5 -> Curr: 20) while text density issues improved; Suggestions: remove watermark-related keywords due to high rejection association.''\\
4. If previous data is missing or empty, base the analysis solely on current high-rejection areas.
\end{promptbox}

\begin{promptbox}{Prompt for Critic-Guided Query Update}
INPUT\\

pending\_subtopics:\\
\{pending\_subtopics\}\\

top\_queries\_examples:\\
\{top\_queries\}\\

recent\_feedback:\\
\{recent\_feedback\}\\

TASK\\

Generate 30--45 Bing image queries for the pending subtopics.\\

STRICT RULES\\

1. Every query must contain at least one text keyword from: readable text, clear text, with text, label, annotation, dimensions, legend.\\
2. Every query must contain at least one quality keyword from: high resolution, 4k, HD, clean layout, no watermark, high contrast.\\
3. Apply the strategy in recent\_feedback. If it suggests avoiding specific patterns or strengthening certain terms, adjust your queries accordingly.\\
4. You are forbidden to copy any domain-specific noun, such as nutrition, packaging, food, etc., from top\_queries\_examples.\\
5. Use the sentence structure style of top\_queries\_examples, but adapt it to the specific domain of pending\_subtopics.\\
6. Output only this JSON, nothing else:\\

\{\\
\hspace*{1em}"queries": ["query1", "query2", "..."]\\
\}
\end{promptbox}

\begin{promptbox}{Prompt for Critic-Guided Generation Prompt Refinement}
You are a prompt rewriter for a text-to-image pipeline that generates images containing visible, readable text, such as signs, labels, documents, posters, and other text-rich visual scenes.\\

A generated image failed verification. Rewrite the prompt to fix the failure.\\

INPUT\\

Original Prompt: \{original\_prompt\}\\
Topic: \{topic\}\\
Subtopic: \{subtopic\}\\
Failure Stage: \{failure\_stage\}\\
Failure Reasons: \{failure\_reasons\}\\
Quality Metrics: \{quality\_details\}\\
Semantic Scores: \{semantic\_details\}\\
Text Consistency Scores: \{text\_details\}\\

RULES\\

Read the Failure Stage and Failure Reasons, then apply the matching rule:\\

- generation failure: Shorten the prompt to under 15 words. Keep only the main subject. Do not include quality words.\\

- quality failure: Keep the original subject. Add: ``crisp text, sharp focus, high contrast, clean white background''.\\

- semantic failure: Start the prompt with the exact topic and subtopic as the subject. Describe the key visual element in one short sentence.\\

- text\_consistency failure: Add this phrase exactly: ``with clearly visible printed text showing '[subtopic]'''.\\

ALWAYS REQUIRED\\

The image must contain readable text. Always include one of the following phrases in the final prompt: ``clear printed text'', ``legible label'', ``readable sign'', or ``visible text overlay''.\\

OUTPUT\\

Return only this JSON. Do not include explanations or extra text:\\

\{\\
\hspace*{1em}"optimized\_prompt": "your rewritten prompt"\\
\}
\end{promptbox}

\begin{promptbox}{Prompt for Experience Library Update}
You are the Experience Library Manager for DataEvolver. Your only job is to maintain a high-quality, traceable experience library that continuously improves data generation quality.\\

INPUT\\

Current semantic feedback:\\
\{current\_feedback\}\\

Recent semantic feedback entries:\\
\{recent\_feedback\_entries\}\\

Top historical queries:\\
\{top\_queries\}\\

K:\\
\{K\}\\

TASK\\

Decide exactly one operation: Add | Delete | Keep\\

- Add: only if the current feedback is genuinely novel and significantly stronger than existing ones.\\
- Delete: only if you detect clear duplication or quality degradation in top\_queries.\\
- Keep: default safe choice when uncertain.\\

STRICT OUTPUT RULES\\

1. You are forbidden to invent, imagine, or recall any feedback, query, or reason not explicitly present in the input above.\\
2. updated\_recent\_feedback must be exactly the last K items, chronological, newest last, and length <= K. Never reorder or drop entries without an explicit Delete operation.\\
3. Every entry in updated\_top\_queries must contain both query and reason.\\
4. When modifying or deleting, keep the original query string traceable.\\
5. Output must be a single valid JSON object and nothing else:\\

\{\\
\hspace*{1em}"operation": "Add|Delete|Keep",\\
\hspace*{1em}"updated\_recent\_feedback": ["...", "..."],\\
\hspace*{1em}"updated\_top\_queries": [\{"query": "...", "reason": "..."\}]\\
\}
\end{promptbox}

\clearpage

%%%%%%%%%%%%%%%%%%%%%%%%%%%%%%%%%%%%%%%%%%%%%%%%%%%%%%%%%%%%
\section{Examples of Rejected Cases}
\label{app:rejected_cases}

This appendix presents representative rejected cases observed during data construction.
These samples are not used for downstream training, but their rejection patterns are summarized as feedback for subsequent construction rounds.

\begin{figure}[htbp]
\centering
\includegraphics[width=0.70\linewidth,height=0.42\textheight,keepaspectratio]{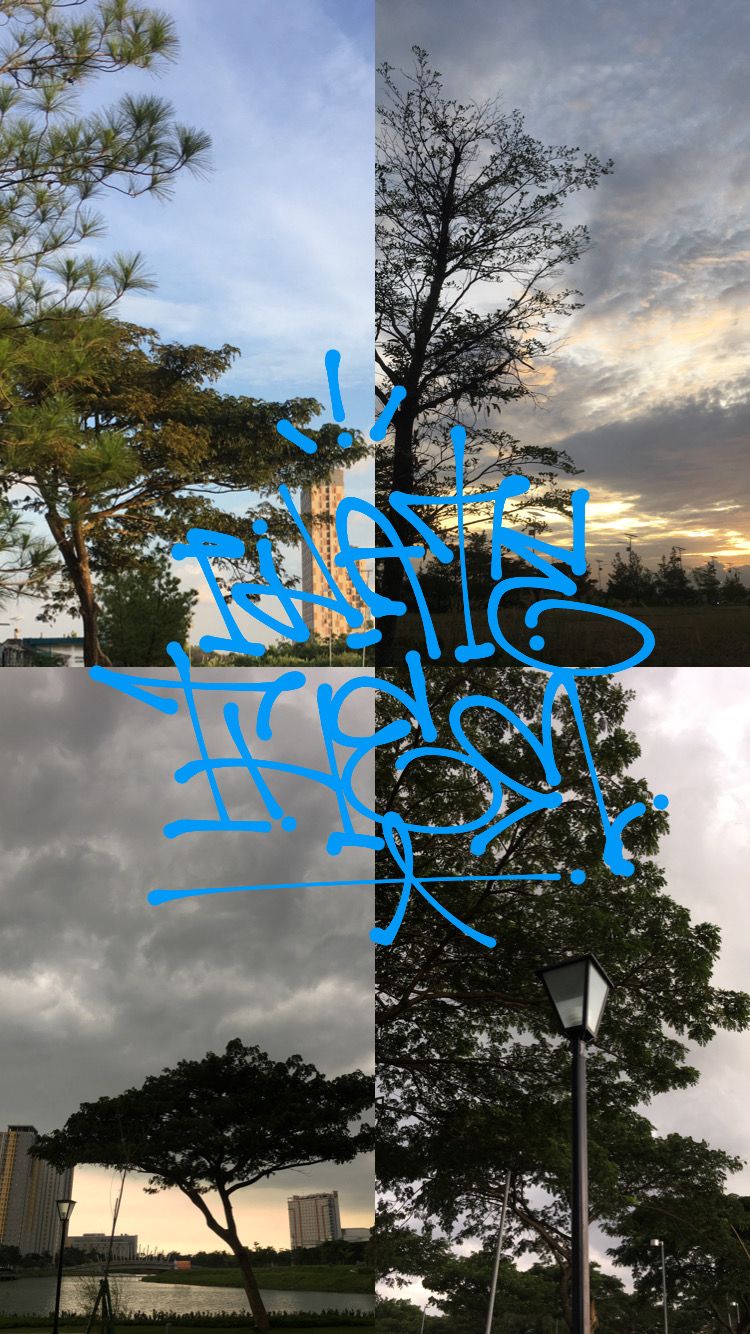}
\caption{
\textbf{Unreadable overlaid text.}
The text is heavily distorted and difficult to recognize reliably.
}
\label{fig:rejected_unreadable_text}
\end{figure}

\begin{figure}[htbp]
\centering
\includegraphics[width=0.70\linewidth,height=0.42\textheight,keepaspectratio]{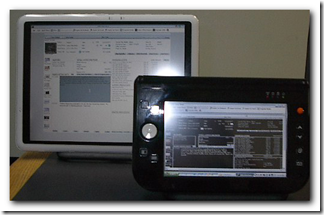}
\caption{
\textbf{Tiny screen text.}
The text occupies only a small region and provides weak OCR supervision.
}
\label{fig:rejected_tiny_screen_text}
\end{figure}

\begin{figure}[htbp]
\centering
\includegraphics[width=0.70\linewidth,height=0.32\textheight,keepaspectratio]{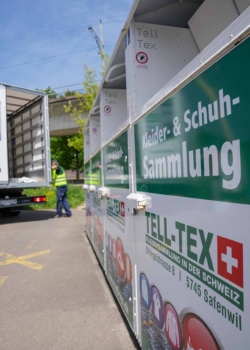}
\caption{
\textbf{Perspective distortion.}
The scene text is affected by viewpoint distortion and background clutter.
}
\label{fig:rejected_perspective_distortion}
\end{figure}

\begin{figure}[htbp]
\centering
\includegraphics[width=0.75\linewidth,height=0.32\textheight,keepaspectratio]{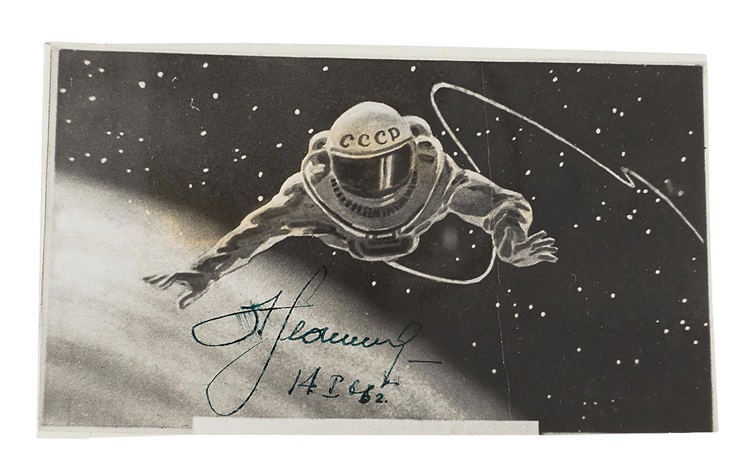}
\caption{
\textbf{Sparse handwriting.}
The image contains limited handwritten text that is difficult to verify consistently.
}
\label{fig:rejected_sparse_handwriting}
\end{figure}

\begin{figure}[htbp]
\centering
\includegraphics[width=0.75\linewidth,height=0.22\textheight,keepaspectratio]{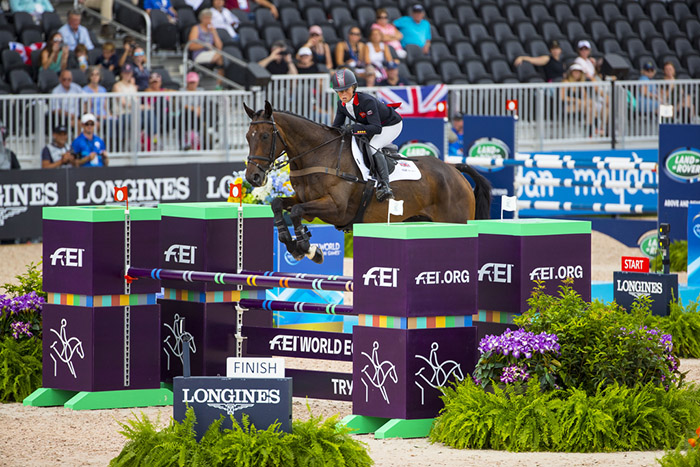}
\caption{
\textbf{Cluttered scene text.}
The image contains complex visual elements and sparse useful text regions.
}
\label{fig:rejected_cluttered_scene_text}
\end{figure}

%%%%%%%%%%%%%%%%%%%%%%%%%%%%%%%%%%%%%%%%%%%%%%%%%%%%%%%%%%%%

\end{document}